\documentclass[12pt]{article}
\usepackage{times}
\usepackage{graphicx} 
\usepackage{subfigure}

\usepackage{natbib}

\usepackage{algorithm}
\usepackage{algorithmic}

\usepackage{hyperref}
\usepackage{rotating}
\usepackage{amsmath}
\usepackage{amsthm}
\usepackage{amssymb}
\usepackage{graphicx}
\usepackage{fullpage}
\usepackage{setspace}
\usepackage{natbib}
\usepackage{color}

\usepackage[T1]{fontenc}
\usepackage{microtype}
\usepackage{amsfonts,amssymb,amsbsy,amsmath,amsthm}
\usepackage{mathtools}
\usepackage{booktabs}

\usepackage[inline]{enumitem}
\usepackage{graphicx,tikz,float}
\usepackage{longtable,relsize}
\usepackage[section]{placeins}

\usepackage{authblk}
\usepackage{hyperref}
\usepackage[margin={1in,1in},includefoot]{geometry}
\usepackage[capitalize,sort]{cleveref}

\usepackage[normalem]{ulem}

\usepackage{array}
\newcolumntype{H}{>{\setbox0=\hbox\bgroup}c<{\egroup}@{}}

\usetikzlibrary{external,arrows,positioning,decorations.pathreplacing}
\hypersetup{breaklinks,colorlinks,allcolors=blue,pdfpagemode=UseNone}

\setlist[itemize]{leftmargin=*}

\newlist{assume}{enumerate}{1}
\setlist[assume]{label=\textup{A\arabic*},leftmargin=*,resume=assume}
\crefname{assumei}{}{}

\newlist{assumer}{enumerate}{1}
\setlist[assumer]{label=\textup{R\arabic*},leftmargin=*,resume=assumer}
\crefname{assumeri}{}{}

\newlist{assumec}{enumerate}{1}
\setlist[assumec]{label=\textup{C\arabic*},leftmargin=*,resume=assumec}
\crefname{assumeci}{}{}

\newlist{assumew}{enumerate}{1}
\setlist[assumew]{label=\textup{W\arabic*},leftmargin=*,resume=assumew}
\crefname{assumewi}{}{}

\Crefname{appendix}{Supplement}{Supplements}
\Crefname{page}{page}{pages}
\crefname{equation}{}{}
\crefname{enumi}{}{}

\makeatletter
\DeclareRobustCommand\citepos
  {\begingroup
   \def\NAT@nmfmt##1{{\NAT@up##1's}}%
   \NAT@swafalse\let\NAT@ctype\z@\NAT@partrue
   \@ifstar{\NAT@fulltrue\NAT@citetp}{\NAT@fullfalse\NAT@citetp}}
\g@addto@macro\@floatboxreset{\small\centering}
\newcommand{\tr}{{\mathpalette\@tr{}}}
\newcommand{\@tr}[2]{\raisebox{\depth}{$\m@th#1\mathsmaller\intercal$}}
\makeatother

\allowdisplaybreaks
\newcommand{\pn}{\mathbb{P}_n}
\newcommand{\ptheta}{{\pi_\theta}}

\title{A Batch, Off-Policy Actor--Critic Algorithm for Optimizing the Average Reward}
\date{}

\author[1]{S.A. Murphy}
\author[2]{Y. Deng}
\author[3]{E.B. Laber}
\author[4]{H.R. Maei}
\author[5]{R.S. Sutton}
\author[6]{K.  Witkiewitz}

\affil[1,2]{University of Michigan}
\affil[3]{North Carolina State University}
\affil[5]{University of Alberta}
\affil[6]{University of New Mexico}

\begin{document}
\maketitle

\begin{abstract}
We develop an off-policy actor--critic algorithm for learning an optimal policy    from a training set composed of data from multiple
individuals.  This algorithm is developed with a view toward its use
in mobile health.  
\end{abstract}

\newpage
\section{Mobile Health}
\label{mhealth}

In the behavioral health communities there is increasing interest in,
and use of, mobile devices to deliver treatments that target behavior
change. Mobile devices can be used to provide treatment when, where,
and in the amount desired (\citealp{Litvin2013, Kumar2013}).
Increasingly scientists are looking to passive sensing (wearable
devices, GPS, activity on the smartphone) and self-report of internal
states to individualize the intervention to the person in terms of
when, how and where to deliver treatment.  Examples of internal states
include level of craving and the perceived need for assistance.  Thus,
scientists are developing treatment policies that encode, on the basis
of the passive and active sensor measures, sequential decisions
regarding when, how and where to deliver treatment. These treatment
policies, also known as, Just-in-Time Adaptive Interventions
\citep{Sprijt2014}, dynamic tailoring \citep{Kennedy2012}, and
intelligent real-time therapy \citep{Kelly2012}, are being used to
intervene on physical activity \citep{King2013}, eating disorders
\citep{Bauer2012}, alcohol use \citep{Witkiewitz2014, Gustafson2014},
mental illness \citep{Depp2010, Ben2013}, obesity/weight management
\citep{Patrick2009} and other chronic disorders \citep{Granholm2012,
  Kristjnsdttir2013}.  In these applications, and throughout much of
mobile health, the treatment policies, e.g., decision rules that input
these measures and output when, how and which treatment to deliver are
formulated using domain expertise.

The main contribution of this paper is the development of an
off-policy, batch, actor--critic algorithm for use in learning
treatment policies from a training set composed of data from multiple
individuals. Actor--critic algorithms have a long history in
sequential decision making \citep{Barto1983, Grondman2012} primarily in
the on-policy, online, setting. These algorithms have been used in
health, for example for on-policy, online glucose regulation in Type 1
diabetes \citep{Daskalaki2013}.  The first off-policy, online
actor--critic algorithm was developed by \cite{Degris2012} and has
been deployed in robotic demonstrations \citep{Gordon2014}.  All of
these algorithms are designed to learn using one long sequence of
interactions.  In contrast, \cite{Silver2013} developed a on-policy,
online temporal-difference algorithm for learning a policy based on
sequences of interactions with multiple individuals.  Here too, we
learn a policy from data from multiple individuals; to our knowledge,
this paper is the first off-policy, batch, actor--critic algorithm for
such use.  We develop this algorithm with a view towards mobile
health.
We will provide a first evaluation of the algorithm via a
series of experiments and illustrate its use with data from a
smartphone study aimed at reducing heavy drinking and smoking.
Here we consider
learning treatment policies that maximize the average reward.

\subsection{Markov Decision Process and Average Reward}

Consider a Markov decision process (MDP), with finite state space
$\cal S$ and finite action space, $\cal A$.  At time $t$,
let $S_t$ be the random
variable denoting the state, $A_t$ be the action,
and $ R_{t+1}$ be the reward.  The probability of the MDP
transitioning to state $s^\prime$ from state $s$ under
action $a$ is
$p\left(s^\prime|s,a\right)=P\left\lbrace
S_{t+1}=s^\prime| S_t=s, A_t=a\right\rbrace$. The expected
reward given that the system occupies state $s$ and action $a$ is
taken is $r(s,a)=E\left(R_{t+1}|S_t=s, A_t=a\right)$; we assume
$r(s,a)$ is bounded over all state, action pairs. We use $\pi$ to
denote a generic stationary policy; $\pi(a|s)$ is the probability that
$A_t=a$ given $S_t=s$ under policy $\pi$.  Throughout we assume that,
for all policies considered, the Markov decision process is
irreducible and aperiodic.
   Let $d^\pi(s)$ denote the stationary
probability of the Markov chain, $S_0, S_1,\ldots$, being in state $s$
under policy $\pi$.  Let $E_\pi$ denote the expectation of
$(S_t,A_t,S_{t+1}, R_{t+1})$ under the steady state distribution,
$d^\pi$.

The average reward is given by,
\begin{eqnarray*}
\eta^\pi&=&\lim_{n\to\infty}{(1/n) E_{\pi}\left[
\sum_{t=0}^n R_{t+1}\Big|S_0=s_0\right]}\cr
&=& {\sum_s d^{\pi}(s)\sum_a\pi(a|s) r(s,a)}.
\end{eqnarray*}
Under the irreducible assumption the above is independent of $s_0$ \citep{Yu2009}.  The differential value of the state $s$ under policy $\pi$ is
\begin{eqnarray*}
V^\pi(s)=\lim_{n\to\infty}E_{\pi}
\left[
\sum_{t=0}^n \left(R_{t+1}-\eta^\pi\right)\Big|S_0=s
\right].
\end{eqnarray*}
  The  Bellman equation  is given by
\begin{eqnarray*}
V^\pi(s)= \sum_a\pi(a|s)
\left\lbrace r(s,a)-\eta^\pi+\sum_{s^\prime}p(s^\prime|s,a) V^\pi(s^\prime)
\right\rbrace
\end{eqnarray*}
for all states, $s$.  Note that neither $r(s,a)$, nor
$p(s^\prime|s,a)$ depend on $\pi$. Under the irreducible assumption,
$V^\pi$ is the unique solution of this equation up to the addition of
a constant, independent of $s$; the Bellman equation gives
rise to a class of $V^\pi$ differing one from another by a constant. A
consequence of this is that in the critic algorithm to follow we need
only learn one of these versions of $V^\pi$.

Here we aim to learn stochastic treatment policies.  One reason for
this, is due to evidence that some variation in actions help
prevent/retard the development of habituation (user ignores action)
\citep[e.g.,][]{Epstein2009}.  This evidence supporting variation may
be partially due to the fact that the MDP is an approximation to the
underlying complex behavioral processes (indeed some parts of the
state space may be yet unknown)and thus even though theoretically the
optimal policy should be deterministic, variation may have desirable
effects.  Furthermore here we aim to learn low-dimensional parametric
stochastic treatment policies so as to facilitate information exchange
with mobile health scientists.

\subsection{Training Data}

The training data consists of $n$ individuals; we assume that the data for
each individual follows an MDP in which the actions are selected
according to a fixed behavior policy, $\mu(a|s)\in (0,1)$ for all
$a\in\mathcal{A}$, $s\in\mathcal{S}$.  On each individual we observe a
trajectory $D=\{S_0, A_0, R_{1},S_1, A_1, R_{2},\dots, S_T, A_T,
R_{T+1}\}$.  $S_0$ is distributed according to an initial state
distribution, $d_0$.  We assume that the trajectories are independent
across individuals and that they are identically distributed.  Let
$E_\mu$ ($P_\mu$) denote the expectation of (probability concerning)  $\{S_0, A_0, R_{1},S_1, A_1,
R_{2},\dots, S_T, A_T, R_{T+1}\}$ under the behavior policy
$\mu$. Assume that the importance weight ${\pi(a|s)}/{\mu(a|s)}$
takes values in $[0, b]$ for some $b$, finite and positive.  Then, the
Bellman equation implies that the average reward for policy $\pi$ can
be written as
  \begin{eqnarray}
 \label{Poisson2}
 \eta^\pi=E_\mu\left[\rho_t^\pi\left\lbrace
     R_{t+1} -V^\pi(S_t) +V^\pi(S_{t+1})\right\rbrace\right]
\end{eqnarray}
for all $t$, where
$\rho_t^\pi= {\pi(A_t|S_t)}/{\mu(A_t|S_t)}$.

The Bellman equation implies that $V^\pi$ satisfies
\begin{eqnarray}
 \label{DiffValue}
 0=
E_\mu\left[\rho_t^\pi \left(
R_{t+1}- \eta^\pi +V^\pi(S_{t+1})-V^\pi(S_t)\right)f_t\right]
\end{eqnarray}
for  $f_t=f(S_t)$ for any $f$, a vector of bounded functions of state and  for all $t$.

\section{Batch, Off-Policy Actor--Critic Algorithm}
Consider a class of parameterized policies, $\pi_\theta(a|s)$ for
$\theta\in {\mathbb{R}^q}$.  The policies  are differentiable in $\theta$ at all states,
actions $(s,a)$.  For example in the experiments below, $\pi_\theta(a|s)=
\frac{e^{-\theta^T\phi(s,a)}}{\sum_{a'}e^{-\theta^T\phi(s,a')}}$,
where $\phi(s,a)$ is a $q$ by $1$
vector of features of the state and action and the parameter $\theta$ indexes the class of policies.  In the mobile health
settings we envision, there will be only a small number of possible
actions and for interpretability, the dimension $q$ will likely be
small as well.

We aim to learn the value of $\theta$ that maximizes the
average reward subject to stochasticity constraints.  In particular, for at least $(1-\alpha)$\% of the states, the probability of  selecting an action in a given state should be at least $p_0$ probability and no more than $1-p_0$ probability.   This goal is operationalized here  by aiming to learn
$\arg\max_{\theta\in\Theta}\, \eta^{\pi_\theta}$ subject to
$ 1-\alpha \le T^{-1}\sum_{t=1}^{T}
P_{\mu}\left[
 p_0 \le \pi_\theta(a|S_t) \le 1-p_0, \ \forall a
\right] \ge 1-\alpha$.

In the following
actor--critic algorithm, the actor algorithm improves the parameters of the policy
resulting in an updated policy. The critic algorithm learns an off-policy
estimate of both the differential value function and the average
reward for the updated policy. The estimated differential value and
estimated average reward are then used by the actor algorithm to again update the
policy.
In the next section, we develop and discuss the
critic algorithm.  In the subsequent section, we develop the
actor algorithm and combine the two  in Algorithm~\ref{alg:ac}.

\subsection{The Critic Algorithm}

Here we discuss off-policy, batch learning of the average reward and
differential value function for a given policy, $\pi$. We consider
linear approximations for $V^\pi(s)$, specifically $v^Tf(s)$ for $f$ a
$p\times 1$ vector of bounded features of the state.
Thus, an  algorithm  for learning the average reward, $\eta$,
and the parameters indexing the differential value, $v$,
may be based on (\ref{DiffValue}) and (\ref{Poisson2}):
\begin{eqnarray}
\label{fixedpt0}
0&=&E_\mu\left[ \rho_t^\pi \left(
R_{t+1} -\eta +v^Tf_{t+1}-v^Tf_t\right)\right]\cr
0&=&E_\mu\left[\rho_t^\pi \left(
R_{t+1}- \eta +v^Tf_{t+1}-v^Tf_{t}\right)f_t\right],
\end{eqnarray}
where $f_t=f(S_t)$.  Note that these equations only involve the value
of the feature vector $f(s)$ up to an additive constant.  Here we
consider feature vectors centered by their empirical mean, e.g., the
feature vectors are constrained to satisfy $\sum_{t=0}^T \pn [f_t]=0$
where $\pn [f_t]=1/n\sum_{i=1}^n f(S_{it})$ and $S_{it}$ is the ith
individual's state at time $t$.

   Define $z_t= (1, f^T_t)^T$
and $\delta_t(\eta,v) =R_{t+1}- \eta +v^Tf_{t+1}-v^Tf_t$.  With this notation (\ref{fixedpt0}) can be written as
\begin{eqnarray}
\label{fixedpt}
E_\mu\left[\rho_t^\pi \delta_t(\eta,v)z_t\right]=0
\end{eqnarray}
for all $t$.  Recall $D=\{S_0, A_0, R_{1},S_1, A_1, R_{2},\allowbreak
\dots, S_T, A_T, R_{T+1}\}$.  An empirical version of (\ref{fixedpt})
is
\begin{eqnarray}
\label{empfixedpt}
\pn\left[\sum_{t=0}^T\rho_t^\pi \delta_t(\eta,v)z_t\right] =0,
\end{eqnarray}
where $\pn [g(D)]=1/n\sum_{i=1}^n g(D_i)$ in which  $D_i$ represents the ith individual's trajectory of states and actions.
The above can, in turn,  be written as
\begin{eqnarray*}
\hat b^\pi - \hat A^\pi \left({ \eta \atop v}\right)=0
\end{eqnarray*}
where $\hat A^\pi= \pn\left[\sum_{t=0}^T\rho_t^\pi \left({ 1 \atop f_t}\right)\left({ 1 \atop f_t-f_{t+1}}\right)^T \right]$
and
$\hat b^\pi=\pn\left[\sum_{t=0}^T\rho_t^\pi \left({ 1 \atop f_t}\right) R_{t+1}\right].$

We use penalization to
control the overfitting due to the high-dimensionality of the features and to ensure uniqueness of the solution as follows.  Given a tuning parameter,
$\lambda_c\ge 0$, we minimize a \lq\lq penalized" norm of
the empirical version of (\ref{fixedpt}),
\begin{eqnarray}
\label{empfixedptCsaba}
\left|\left|\hat b^\pi-\hat A^\pi\left({ \eta \atop v}\right)\right|\right|_2^2 + \lambda_c ||v||_2^2
\end{eqnarray}
for $\hat\eta$ and $\hat v$.
 The minimizer of this equation satisfies
\begin{eqnarray}
\label{zeros}
\left(\hat A^\pi\right)^T\hat b^\pi &=&
\left\lbrace \left(\hat A^\pi\right)^T\hat A^\pi + \lambda_c
\tilde I_{p+1}\right\rbrace\left({ \hat\eta \atop \hat v}\right)
 \end{eqnarray}
 where
$\tilde I_{p+1}= \left(\begin{smallmatrix}
0& 0_p^T\\ 0_p&I_p
\end{smallmatrix}\right)$.
%
%
In the experiments below we select the tuning parameter $\lambda_c$ by
cross-validation \citep{Hastie2009}.  The critic algorithm is shown in
Algorithm~(\ref{alg:critic}).

\begin{algorithm}[tb]
   \caption{Critic Algorithm}
   \label{alg:critic}
\begin{algorithmic}
   \STATE {\bfseries Input:} $\theta, D=\{D_i,\ i=1,\ldots,n\}, f, \mu$
   \STATE  $\pi=\pi_\theta$
   \STATE  $\rho=\pi/\mu$
   \STATE Select $\lambda_c$ to minimize the first term in (6) via k-fold cross validation
    \STATE Calculate $\hat A^\pi,\ \hat b^\pi$ from $D, f, \rho$
    \STATE Solve for $\left({ \eta \atop  v}\right)$ in
   $$\left(\hat A^\pi\right)^T\hat b^\pi =\left\lbrace \left(\hat A^\pi\right)^T\hat A^\pi + \lambda_c
\tilde I_{p+1}\right\rbrace\left({ \eta \atop  v}\right)$$ to obtain $\left({ \hat\eta \atop  \hat v}\right)$
  \STATE {\bfseries Output:} $J(\theta)=\hat\eta$
\end{algorithmic}
\end{algorithm}

\subsection{The Actor  and Actor--Critic Algorithms}
Recall from (\ref{Poisson2}) that the average reward under policy
$\ptheta$ is given by $E_\mu\left[\rho^\ptheta_t\left\lbrace R_{t+1}
    +V^\ptheta(S_{t+1})-V^\ptheta(S_t)\right\rbrace\right]$ for all $t$.
The critic algorithm approximates the differential value by a linear
approximation, i.e., $V^\ptheta(s) = v_\theta^Tf(s)$.  Thus a possible
objective function for the actor algorithm is
$$ J(\theta)=\pn\left[\sum_{t=0}^T\rho_t^\ptheta\left(
    R_{t+1} + \hat v_\theta^Tf_{t+1}- \hat
    v_\theta^Tf_t\right)\right]$$ where $\hat v_\theta$ is provided by
the critic.  That is $\hat v_\theta$ is given by $\hat v $ from
(\ref{zeros}) for $\pi=\ptheta$.  See \cite{Maei2013} for a similar
objective function in the off-policy discounted horizon setting.

Because $J(\theta)$ is not
concave and tends to be flat for large entries in $\theta$, we use a
quadratic penalty to stabilize the optimization, namely
\begin{equation}\label{penJtheta}
J(\theta) -
\lambda_a \theta^T\Sigma \theta,
\end{equation}
where $\Sigma$ is a  matrix (here we use
$\Sigma=\pn \sum_{t=1}^{T}\phi(S_t, A_t)\phi^{T}(S_t, A_t)$)
and $\lambda_a\ge 0$ is a tuning parameter.  This penalty
shrinks $\theta$ toward zero so that the estimated policy
is shrunk toward a uniform policy over actions.  The tuning parameter on the penalty, $\lambda_a$, is selected to ensure that the learned
treatment policy will, for $(1-\alpha)$\% of the states, select each action with at
least $p_0$ probability, e.g., $0.05$, and no more than $1-p_0$
probability.

The actor algorithm performs the maximization of (\ref{penJtheta}) over $\theta\in \mathbb{R}^{q}$.
In the experiments and the data example below, the solution to the maximization is computed  using the Broyden-Fletcher-Goldfarb-Shanno (BFGS)
algorithm with multiple random starting values to avoid local maxima.
We use the implementation of BFGS with a finite-difference approximation
to the gradient in the \texttt{optim} function of
\texttt{R} (https://stat.ethz.ch/R-manual/R-devel/library/stats/html/optim.html).  The BFGS algorithm is iterative, repeatedly calling the critic algorithm to obtain $J(\theta)$ for different values of $\theta$.

The actor algorithm is represented by the maximization steps in the  batch, off-policy, actor--critic algorithm  given in Algorithm~\ref{alg:ac}.  Note that if $p_0$ is less than $1/K$ where $K$ is  the number of possible actions and $\Sigma$ is full rank then the while loop will terminate.  This will occur because  a very large tuning parameter, $\lambda_a$ will lead to a solution for $\hat\theta$ that is close to a vector of $0$'s and in this case $\pi_{\hat\theta}(a|s)$ is approximately equal to $1/K$  for all states, $s$ and actions $a$.

\begin{algorithm}[tb]
  \caption{Batch, Off-Policy, Actor--Critic Algorithm}
  \label{alg:ac}
  \begin{algorithmic}
    \STATE {\bfseries Input:} $D,\,f,\,\mu,\,\Sigma,\, p_{0},\,
    \lambda_{a}^{\mathrm{min}},\, \Delta>0$
    \STATE $\lambda_a=\lambda_a^{min}$
    \STATE {\bfseries Actor Step:} ${\hat\theta} = \arg\max_{\theta}\left\lbrace
      J(\theta) - \lambda_{a}\theta^{T}\Sigma \theta
      \right\rbrace$
\WHILE{$\min_{a}T^{-1}\sum_{t=1}^{T} \pn \left[1_{ p_{0} \le \pi_{\hat\theta}(a|S_t)\le 1-p_{0}}\right] < 1-\alpha$}
   \STATE $\lambda_a=\lambda_a+\Delta$
\STATE {\bfseries Actor Step:} ${\hat\theta} = \arg\max_{\theta}\left\lbrace
      J(\theta) - \lambda_{a}\theta^{T}\Sigma \theta
      \right\rbrace$
      \ENDWHILE
\STATE {\bfseries Output:} $\pi_{\hat\theta}$
  \end{algorithmic}
\end{algorithm}


\section{Experiments}
We first use a series of experiments to examine the finite sample
performance of the actor--critic algorithm. Second we apply the
actor--critic algorithm to a data set concerning a mobile health
intervention for college students who drink heavily and smoke
cigarettes \citep{Witkiewitz2014}.  In both cases the actions are
binary, coded to take values in $\lbrace 0, 1\rbrace$.  Here, $a=1$
means providing the active treatment, e.g., sending an intervention to
the subject's mobile device, and $a=0$ means no treatment.  We
restrict to classes of policies of the form
$\pi_{\theta}(1|s) = \left\lbrace
  1+\exp(\theta^{T}\phi(s))\right\rbrace^{-1}$
where $\theta \in \mathbb{R}^q$.  In these experiments, the critic
algorithm uses $2$-fold cross-validation (other choices of the number
of folds are possible: \citep{Hastie2009}) to select
$\lambda_{c}$. 

\subsection{Simulated Experiments}
 Define
$\pi_{\mathrm{opt}}$ to be the
solution to
$\max_{\theta\in\mathbb{R}^q}\, \eta^{\pi_\theta}$ subject to $
T^{-1}\sum_{t=1}^{T}P_{\mu}\left[ p_0 \le \pi_\theta(1|S_t) \le 1-p_0
\right] \ge 1-\alpha$.  Throughout we set $p_0=\alpha=.05$.
We use the
performance of $\pi_{\mathrm{opt}}$ as a gold standard in assessing the performance of
the actor--critic algorithm; $\pi_{\mathrm{opt}}$ is computed  by  approximating
$\eta^{\pi_\theta}$ using the generative model and optimizing $\eta^{\pi_\theta}$ using the BFGS algorithm with an exact penalty
to enforce the constraint \citep{bertsekas2014constrained}.
To form a clinically
relevant baseline for comparison we also consider the constant policy
$\pi_{\mathrm{const}}(1|s) \equiv 1$ for all $s$.  The constant policy
aligns with the common perspective that more treatment leads to better
patient outcomes, however, such a policy risks over-burdening the
patient potentially leading to poor average reward.

We compare the proposed actor--critic algorithm,
with the use of the optimal policy, $\pi_{\mathrm{opt}}$, and the constant
policy $\pi_{\mathrm{const}}$ in terms of average reward.  For any
policy $\pi$ we calculate the average reward, $\eta^{\pi}$, using
a large independent test set
 generated using $\pi$.  We measure of the performance of the actor--critic algorithm via
$E_\mu\left[\eta^{\pi_{\widehat{\theta}}}\right]$; note $\eta^{\pi_{\widehat{\theta}}}$ depends on the training data via $\widehat\theta$, thus the expectation, $E_\mu$.

In the experiments the subjects' trajectories
$\{S_0, A_0, R_{1},S_1, A_1, R_{2},\allowbreak \dots, S_T, A_T,
R_{T+1}\}$ are i.i.d. and are simulated as follows. The behavior policy selects the action coded 1 with probability $.6$ throughout (i.e., $\mu(1|s)=.6$ for all states $s$).   The state, $S_t$, is a $p_1\times 1$
vector. For $\rho \in (0,1)$ define $\mathrm{AR}(\rho)$ to be the
$p_1\times p_1$ matrix $(\mathrm{AR}(\rho))_{ij} = \rho^{|i-j|}$.  In the following class of generative models, the evolution of all of the states except the ``burden" state,  $S_{t,3}$, is according to a stochastic linear system.  The burden, $S_{t,3}$,  is generated so that when treated, the  burden
increases approximately linearly with slope $0.5$ and when not
  treated, the treatment burden decreases approximately
  geometrically with rate $0.90$; in particular  $E[S_{t,3}|S_{t-1,3}=s_3,
  A_{t-1}=1] = 0.95s_3 + 0.5$ whereas $E[S_{t,3}|S_{t-1,3}=s_3,
A_{t-1}=0] = 0.9s_3$.
The initial state and action are generated by $    \mathbf{S}_{0} \sim \mathrm{Normal}_{p_1}\left\lbrace
      0, \mathrm{AR}(0.5)\right\rbrace$  and for $t\ge 1$,
  \begin{eqnarray*}
    \xi_t &\sim & \mathrm{Normal}_{p_1+1}(0, I), \\
    U_t &\sim & \mathrm{Uniform}[0,1]^2,\\
    S_{t,1} &=& 0.5 S_{t-1,1} + 2\xi_{t,1 },\\
    S_{t,2} &=& 0.25 S_{t-1,2}+ 0.125 A_{t-1} + 2\xi_{t,2}, \\
    S_{t,3} &=& 0.9S_{t-1,3} + 0.1S_{t-1,3}U_{t,1}A_{t-1} + U_{t,2}A_{t-1}, \\
    S_{t,j} &=& 0.25S_{t-1,j} + \xi_{j},\,j =4,\ldots, p_1, \\
    R_{t+1} &=& 10 + 0.25S_{t,1}
    A_{t}(0.04 + 0.02S_{t,1}+0.02S_{t,2})
    \\ && - \tau S_{t,3} + 0.16\xi_{t,p_1+1}.
  \end{eqnarray*}
  This class of models is indexed by the number of state variables, $p_1$ and $\tau$ where $\tau$ represents the  impact of burden  $S_{t,3}$ on the reward. Note that in this generative model, $S_{t,3}$ does not influence how the current action ($A_t$) impacts the reward, but rather leads to an overall reduction in the current reward regardless of the the present action $A_t$.
   With the exception of the
  noise variables, $S_{t,j},\, j\ge 4$, the effect sizes in the
  generative model are loosely based on the BASICS-Mobile
  data presented in the next section.

We use a linear approximation to the differential value of the form
$v^{T}f(s)$.  The feature vector, $f$ is constructed from state $s
\in\mathbb{R}^{p_1}$ using a special case of multivariate adaptive
regression splines \citep{Friedman1991}.  In particular, define $c_{j,k}$,
$k=1,\ldots,10$ to be the sample deciles of the $j$th component of the
state vector.  The feature vector, $f$ consists of the vector of all
singletons and pairwise products of piecewise linear splines in the
set: $\{(s_{j} - c_{j,k})_{+}$, $(c_{j,k}-s_{j})_{+},\
j=1,\ldots,p_1,\ k=1,\ldots,10\}$ (note $(u)_{+}= \max(0,u)$).  Thus,
$p$, the dimension of the feature vector, $f$ is on the order
 $600p_1^2$.
To reduce computation we exclude, from $f$, basis
functions that are zero for more than 80\% of the observed
states in the training set.  
As mentioned previously, these features are centered
to have empirical mean zero as the differential value is
only defined up to an additive constant.

In all simulated experiments we use training sets of $n=25$ individuals observed over
$T=25$ time points. The average reward is calculated for a policy
$\pi$ by averaging the rewards from the last 9,000 elements
 a trajectory of length 10,000 under the policy $\pi$.
Expectations with respect to the distribution underlying the
training set, $E_{\mu}$, are approximated using
$100$ Monte Carlo replicated training sets.

\begin{figure}[ht]
\vskip 0.2in
\begin{center}
\centerline{\includegraphics[width=\columnwidth]{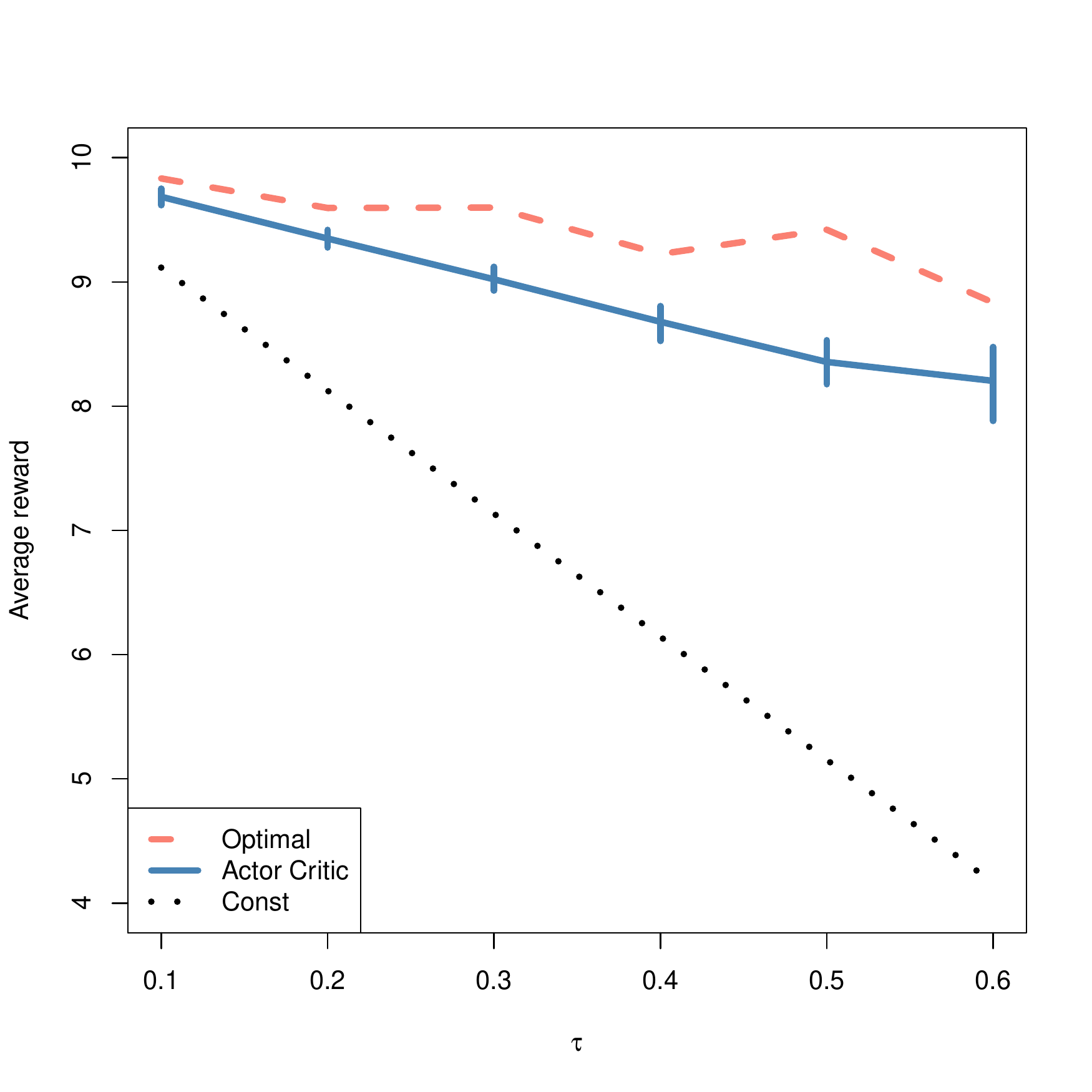}}
\caption{Average reward for simulation setting (S1).  The actor--critic algorithm (solid) tracks the gold standard (dotted).  As burden increases
the performance of the constant policy (dashed line) decreases rapidly.
}
\label{s1}
\end{center}
\vskip -0.2in
\end{figure}

\begin{figure}[ht]
\vskip 0.2in
\begin{center}
\centerline{\includegraphics[width=\columnwidth]{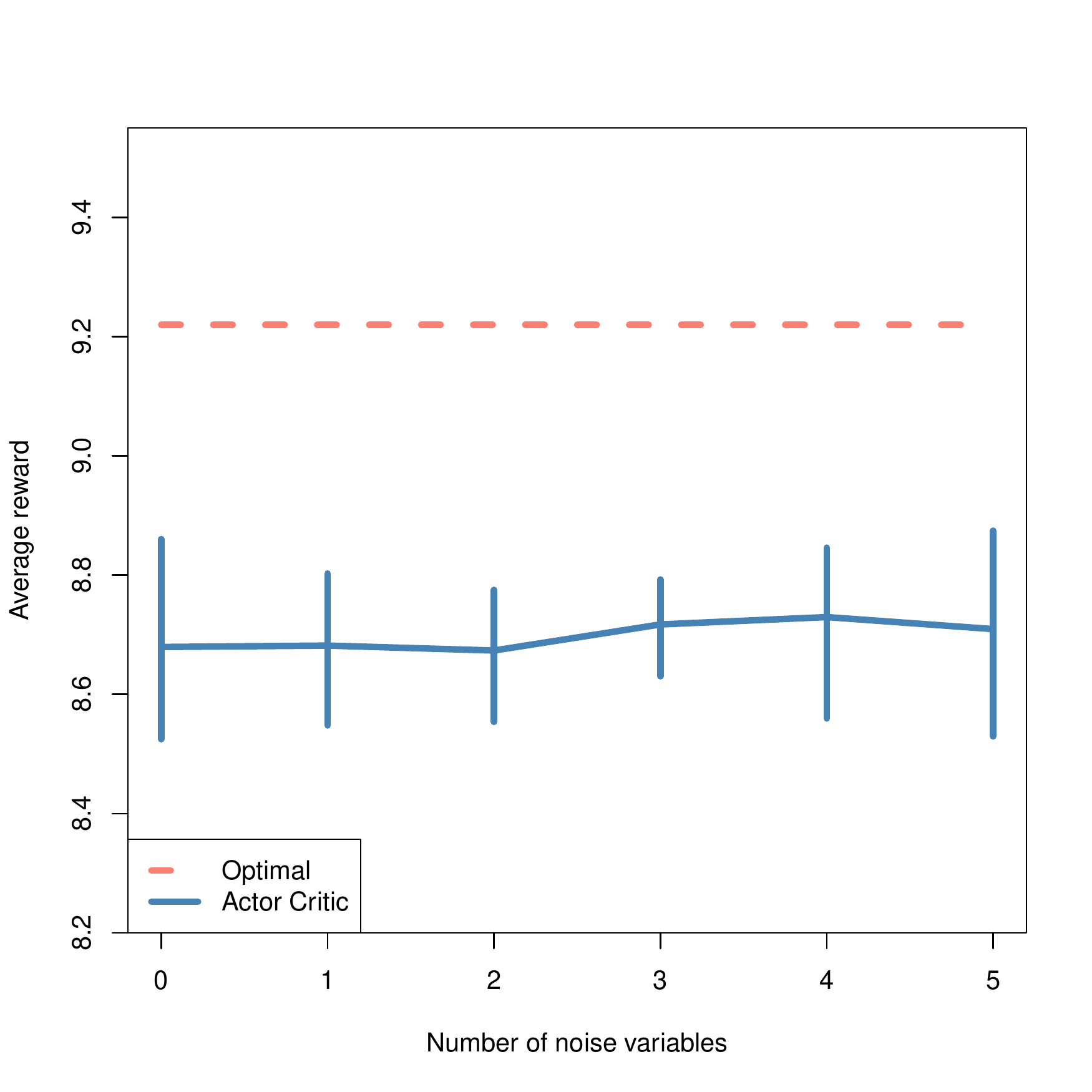}}
\caption{Average for simulation setting (S2); recall that
$\tau=0.4$ is fixed.  As the number
of noise variables in the approximation for the differential
value increases
the performance actor--critic algorithm (solid)
does not appear to deteriorate.  There are no noise variables in the policy.}
\label{s2}
\end{center}
\vskip -0.2in
\end{figure}

\begin{figure}[ht]
\vskip 0.2in
\begin{center}
\centerline{\includegraphics[width=\columnwidth]{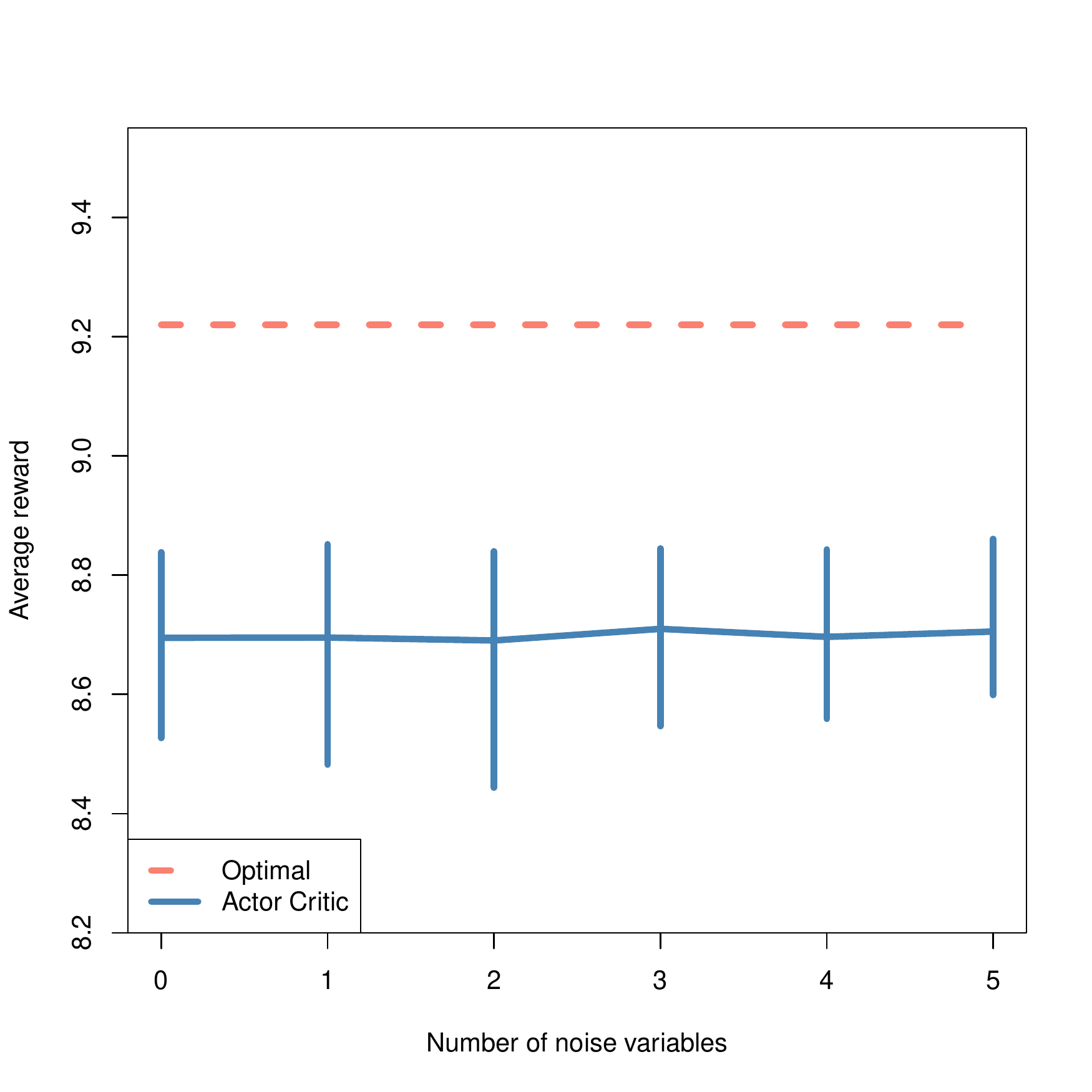}}
\caption{ Average for simulation setting (S3);  recall that
$\tau=0.4$ is fixed. The addition
of noise variables into the policy  as well as in the approximation to the differential value only moderately increases the
variability, across training sets, in the average reward of the learned policy.}
\label{s3}
\end{center}
\vskip -0.2in
\end{figure}

\begin{figure}[ht]
\vskip 0.2in
\begin{center}
\centerline{\includegraphics[width=\columnwidth]{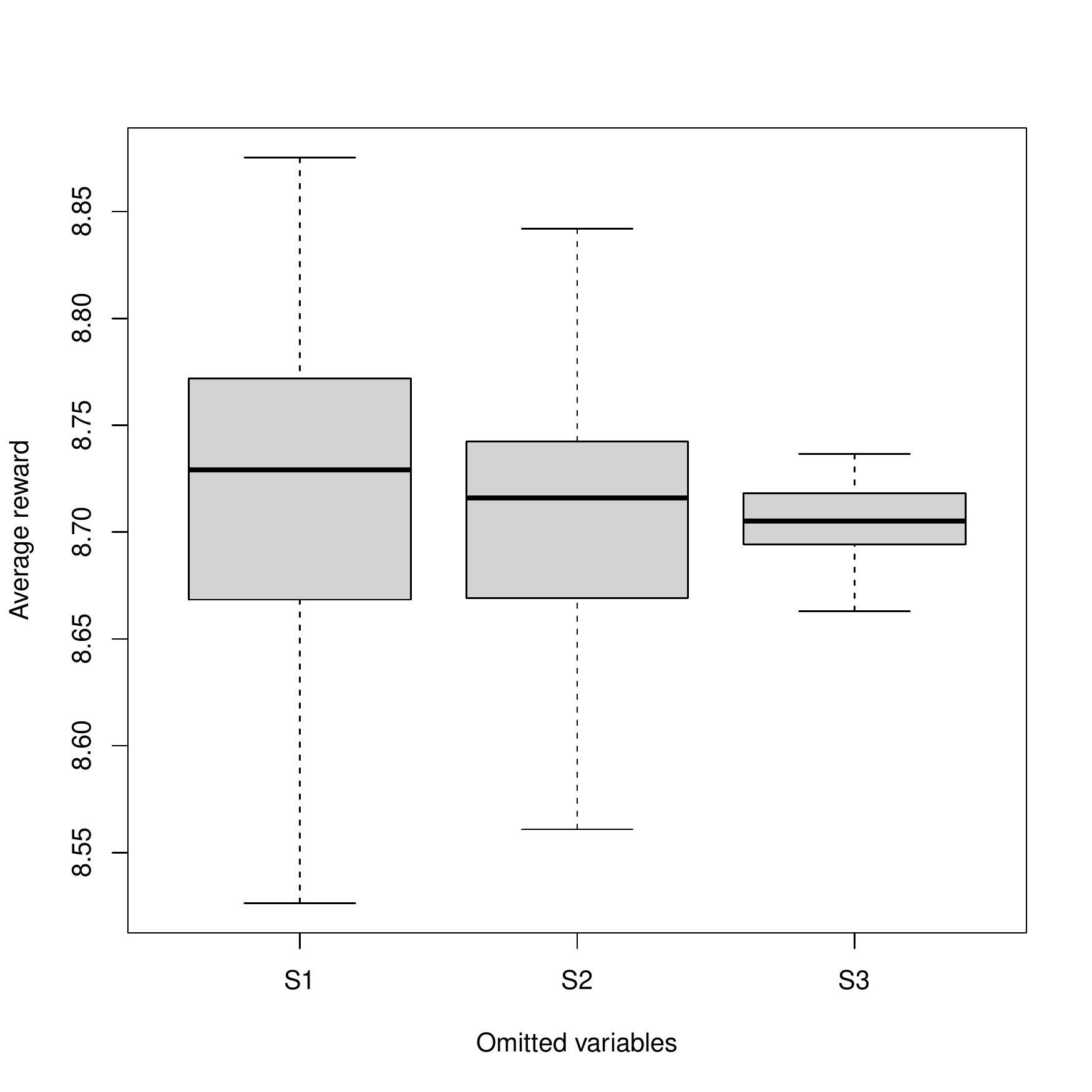}}
\caption{Average for simulation setting (S4); recall that
$\tau=0.4$ is fixed.  Omitting the state
$S_{t,3}$ is associated with a decrease in the median average award as
well as a decrease in the variability, across training sets, of the average reward of the learned policy.}
\label{s4}
\end{center}
\vskip -0.2in
\end{figure}

We consider the following experiments to illustrate different
aspects of the actor--critic algorithm:
\begin{enumerate}
\item[(S1)] In this example, the state vector, $S_t$ has dimension
  $p_1=3$, and each member in the policy class has $q=4$ parameters, an
  intercept plus coefficients of $S_{t,1}, S_{t,2}, S_{t, 3}$.  The
  burden effect parameter, $\tau$ ranges from $0.20$ to $0.60$;
  as $\tau$ moves across this range, the performance of the
  constant policy decreases by approximately 50\%.
  The
  purpose of this example is to assess how well the proposed
  actor--critic algorithm learns a policy with average reward that tracks the
  average reward of the optimal policy  as the effect of burden
  changes.  Also this example  illustrates the effect of burden on
  the constant (always treat) policy relative to the learned and
  optimal policies.  A brief description of how the optimal policy is computed was given at the beginning of Section~3.1. Figure (\ref{s1}) shows $E_\mu\left[\eta^{\pi_{\hat\theta}}\right]$,
the gold standard, and the constant policy. Recall $E_\mu\left[\eta^{\pi_{\hat\theta}}\right]$ is approximated by an average of 100 $\eta^{\pi_{\hat\theta}}$'s learned from 100 Monte-Carlo replications of the training set.   Tick marks indicate the
5th and 95th percentiles over  the 100 $\eta^{\pi_{\hat\theta}}$'s.
In this example the average reward
of the proposed algorithm tracks the gold standard closely while
the constant policy performs poorly, especially as burden increases.

  \item[(S2)] In this example the burden effect $\tau = 0.4$; the
    dimension of the state vector, $S_t$ ranges from $p_1=3$ to
    $p_1=10$, and each member in the policy class has $q=4$ parameters, an
    intercept plus coefficients of $S_{t,1}, S_{t,2}, S_{t, 3}$.
    Thus, in this example, when $p_1 > 3$, there are additional noise
    variables used to approximate the differential value but these variables
    are not used in the policy class.  This example reflects our
    perspective that in mobile health scientists are accustomed to
    identifying important variables that might enter
    policies based on their expertise, but do not have experience in
    identifying important variables for the differential value.  The
    optimal policy is the same as in (S1) for $\tau=0.4$.  Figure (\ref{s2}) provides the results.  The average
reward for the proposed algorithm tracks the gold standard (which is
unaffected by noise variables) and remains stable as the number of
noise variables added to the model increases.

  \item[(S3)] In this example $\tau = 0.4$, $p_1$ ranges from $3$ to
    $10$, and $q={p_1+1}$.  This example represents
    the setting where noise variables are included in  both the policy and the approximation to  the differential value.  The optimal policy is the same
    as in (S2).
  Figure (\ref{s3})  shows that the proposed algorithm
  is relatively robust to noise variables in the policy even
in data-impoverished settings with both $T$ and $n$ small.

  \item[(S4)] In this example $\tau = 0.4$ and we consider the
    setting where $p_1 = 3$ but one  of the
    three state variables, $S_{t,1}$, $S_{t,2}$, $S_{t,3}$, have
    been omitted from the policy ($q=3$).  The optimal policy which includes an intercept and $S_{t,1}$, $S_{t,2}$, $S_{t,3}$ is the same as in (S1) when $\tau=0.4$. Figure (\ref{s4}) illustrates that omitting the state variable, encoding burden,
$S_{t,3}$, generally reduces the median outcome but also reduces variability.  We conjecture that this is due to the fact that in the generative model for $R_{t+1}$, $S_{t,3} $ does not interact with the action, $A_t$.
\end{enumerate}

\subsection{BASICS-Mobile}
BASICS-Mobile is a mobile intervention targeting smoking and heavy
episodic drinking by college students \citep{Witkiewitz2014}.
Mobile interventions are attractive because of their ability to provide feedback about drinking or smoking as the person goes about his/her daily life.
This intervention contained treatment modules targeting drinking as well as smoking.   These modules contained 1-3 mobile phone screens of content and are interactive in that the student answers  brief questions with responses from the system tailored to their answers. Example modules are a module that provides feedback about smoking, comparing the student's smoking level with the smoking levels of similar students and a module that provides strategies to help the student recognize urges to smoke and strategies for managing smoking urges.

An interesting question that arises in this setting is when to provide treatment modules.   The modules can be burdensome often encouraging students to think about something that they might not, at the time, be receptive to thinking about.   Students may be less responsive to a treatment module if for example, they are already feeling burdened by the mobile intervention or if they are feeling depleted, depressed or stressed out in the moment. These latter “self-control demands” include the need to regulate mood, control thoughts or deal with stress and may decrease a student’s willingness to complete the module.  In the following we use the  actor--critic algorithm to learn a proposal for a treatment policy that would pinpoint when to provide treatment modules.

%
%
%

The study enrolled 29 students; we use data from $n=27$ students, omitting data from two students with large amounts of missing data.
All other  missing data was singly
imputed with the fitted value from a local polynomial regression of the
state variable on time $t$.
The study lasted 14 days and on
the afternoon and evening of each day, treatment modules may be provided, thus, there
are $T=28$ time points per student.  To be available to  receive a
treatment module a student must first complete a list of
self-report questions.   For each student and at each time point define the availability indicator, $I_t=1$ if the student is available, that is, the student completes the self-report questions,  and $I_t=0$ otherwise.
Students completed the self-report
questions 86\% of the time in this study and thus were available for a treatment module at 85\% of the time points.
If a student is available
at a time point then the student may be provided a treatment
module ($A_t=1$) or a general informational/health module ($A_t=0$).  So $A_t$ can only
occur at a time $t$, if
$I_t=1$. A treatment module is delivered, i.e., $A_t=1$,  approximately $2/3$ (.68) of the times that $I_t=1$.
Because it is not feasible to provide treatment modules
when $I_t=0$, the class of  policies is of the form $\pi_{\theta}(1|s) =
I_{t}\left\lbrace 1+\exp(\theta^{T}\phi(s))\right\rbrace^{-1}$,
where $\phi(s)$ is a feature vector.

Consistent with the discussion above indicating that it might not be beneficial to provide a treatment module at each time point, the measurements that we include in the policy should  act as proxies for self-control demands and treatment burden.  One of the self-report questions is a measure of self-control demands: ``how much do you feel that you need to control or fix your mood,'' coded as 0-4 (not at all to very much). In the policy we  include the change from the prior time to the current time in self-control demands,  \texttt{deltacontrol}: an
indicator coded by 1 for increase and 0 otherwise.  As a proxy for treatment burden we use past availability; recall availability is coded by 1 if the student completes the self-report questions and by 0 otherwise. A student who is feeling burdened by the intervention may ignore the requests by the mobile device to answer the questions or may stop midway through the questions.  In the policy we include, \texttt{burden}: coded by 1 if $I_{t-1}=1$ and 0 otherwise.

In addition to the above two measurements, the state vector at each
time $t$ consists of a further six measurements: (i)
\texttt{smoke}: the average number of cigarettes smoked per hour since
the last report; (ii) \texttt{pastsmoke}: \texttt{smoke} at the
preceding time point; (iii) \texttt{pasttxt}: an indicator of
treatment at the last time point, e.g., $A_{t-1}$ if $I_t=1$ and $0$
otherwise; (iv) \texttt{urge}: student reported agreement with the
statement ``I have a strong urge for a cigarette now,'' coded 0-4
(strongly disagree to strongly agree) (v) \texttt{pasturge}:
\texttt{urge} at the preceding time point; and lastly (vi) current availability, $I_t$.

The feature vector indexing the policy
contains an intercept, \texttt{deltacontrol} and \texttt{burden} (e.g. $q=3$). In a similar manner to  the simulated experiments, the tuning parameter, $\lambda_c$ is selected so that
$\frac{\sum_{t=1}^{28}\pn\left[ .05 \le \pi_\theta(1|S_t) \le .95 \bigcap I_t=1
\right]}{ \sum_{t=1}^{28}\pn\left[ I_t=1\right]}\ge .95$.
The differential value is approximated by a linear combination of MARS basis functions as in the experiments; all  eight variables are used to construct the basis functions.
The reward is the  negative of
\texttt{smoke} measured subsequent to treatment.

The coefficients indexing the learned  policy
are displayed in Table (\ref{basicsCoefTable}).
Under the learned policy  a student with no increase in self-control demands and who is not indicating burden is recommended treatment with probability 0.75 whereas a student who has experienced an increase in self-control demands and who is indicating burden is recommended treatment with probability .51.  This proposed policy
is consistent with the above discussion that students feeling self-control demands and/or experiencing treatment burden are less receptive to a treatment module and thus delivering a treatment module is less likely to be useful in reducing smoking.

\begin{table}
   \centering
   \caption{\label{basicsCoefTable}
    Coefficients indexing estimated optimal policy for BASICS-Mobile
    data.    A student with no increase in self-control demands and who is not indicating burden is recommended treatment with probability 0.75 whereas a student who has experienced an increase in self-control demands and who is indicating burden is recommended treatment with probability .51.
  }
  \vskip 0.15in
  \begin{small}
  \begin{sc}
  \begin{tabular}{lc}
    \hline
    Variable & $\widehat{\theta}$ \\
	\hline
    Intercept & 0.45 \\
    \texttt{deltacontrol} & -0.42  \\
    \texttt{burden} & 0.63\\
	\hline
  \end{tabular}
  \end{sc}
  \end{small}
  \vskip -0.1in
\end{table}

\section{Discussion and Future Directions}

This work represents a first start toward learning policies from training data sets.  The robustness of results to the use of noise variables in approximating the differential value is promising and indicates that perhaps this approximation can be automated.  Such would reduce the burden and enable  domain scientists to focus on which information should be in the state and which state information should be part of the policy.   Critical generalizations include incorporating baseline data (e.g. gender, severity of disorder, genetics) into the algorithm and providing measures of confidence for the $\theta$ parameters in the policy so that domain scientists can decide whether potentially expensive variables should be collected in order to roll out the policy.  Also measures of confidence would enable domain scientists to test behavioral theories.

\bibliography{cites}

\vspace{0.5in}
\noindent
{\it Acknowledgements:} Funding was provided by the National Institute on Drug Abuse (P50DA039838, R01DA039901, R01DA015697), National Institute on Alcohol Abuse and Alcoholism (R01AA023187), National Heart Lung and Blood Institute (R01HL125440), and National Institute of Biomedical Imaging and Bioengineering (U54EB020404).
\end{document}